\pdfoutput=1
\documentclass[11pt]{article}

\usepackage[preprint]{acl}

\usepackage{amsmath}
\usepackage{times}
\usepackage{latexsym}
\usepackage{array} % required for text wrapping in tables
\usepackage[T1]{fontenc}
\usepackage[utf8]{inputenc}

\usepackage{microtype}
\usepackage{graphicx} % DO NOT CHANGE THIS
\usepackage{booktabs}
\usepackage{multirow}
\usepackage{inconsolata}
\usepackage{booktabs}
\usepackage{algorithm}
\usepackage{algorithmic}
\usepackage{comment}
\usepackage{xcolor}
\usepackage{array}
\newcommand{\model}{\textsc{GuideQ}}

\title{\model: Framework for Guided Questioning for progressive informational collection and classification}

\author{Priya Mishra \textsuperscript{\textbf{$\boldsymbol{\ddag}$}} \hspace{1em} Suraj Racha \textsuperscript{\textbf{$\boldsymbol{\ddag}$}} \hspace{1em}  Kaustubh Ponkshe\\  {\bf Adit Akarsh} \hspace{1em}  {\bf Ganesh Ramakrishnan} \\
        Indian Institute of Technology Bombay}
\begin{document}
\maketitle
\def\thefootnote{\textbf{$\boldsymbol{\ddag}$}}\footnotetext{Equal contributions}\def\thefootnote{\arabic{footnote}}
\begin{abstract}
Question Answering (QA) is an important part of tasks like text classification through information gathering. These are finding increasing use in sectors like healthcare, customer support, legal services, {\em etc.}, to collect and classify responses into actionable categories. LLMs, although can support QA systems, they face a significant challenge of insufficient or missing information for classification. Although LLMs excel in reasoning, the models rely on their parametric knowledge to answer. However, questioning the user requires domain-specific information aiding to collect accurate information. Our work, \model, presents a novel framework for asking guided questions to further progress a partial information. We leverage the explainability derived from the classifier model for along with LLMs for asking guided questions to further enhance the information. This further information helps in more accurate classification of a text.
\model{} derives the most significant key-words representative of a label using occlusions. We develop \model's prompting strategy for guided questions based on the top-3 classifier label outputs and the significant words, to seek specific and relevant information, and classify in a targeted manner. Through our experimental results, we demonstrate that \model{} outperforms other LLM-based baselines, yielding improved F1-Score through the accurate collection of relevant further information. We perform various analytical studies and also report better question quality compared to our method.
\footnote{Code available at: \url{https://github.com/SDRMp/DRPG}}
%The applications of Large Language Models (LLMs) have seen an increasing surge guided chat completions especially when the input text is partial and incomplete. Although LLMs have been remarkable in various NLP tasks, their performance on asking guided questions for a domain specific task is limited. This is largely due to the fact that narrowed down tasks requires query specific information which the model lacks. Our work, \model, presents a framework for classification using guided clarification questioning specific to the query. Our framework leverages the LLM using post hoc explanations of the most probable classifier along with the most representative key words of the label learnt during classifier training to ask a personalized information seeking question. We show that grounded response to the generated question proves as more complete by increase in the classification accuracy of the text. We also show that \model{} generates more precise and relevant guided clarification questions as compared to other LLM-based baselines.

\end{abstract}

\begin{figure}
    \centering
    \includegraphics[width=1\linewidth]{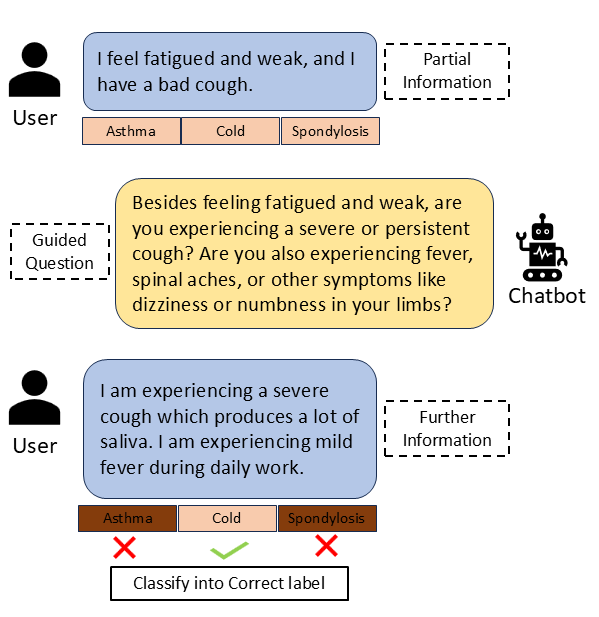}
    \caption{Illustration of partial information by user followed by a specific guided question}
    \label{fig:intro_diag}
\end{figure}

\section{Introduction}
Question Answering (QA) systems have been an integral part of the NLP landscape \cite{QA_gen}. In particular, the emergence of LLMs has enabled reasoning, proactive questioning, and better semantic understanding of the user response during questing answering or dialogue~\cite{QA_LLM}. 
\begin{figure*}[h]
    \centering
    \includegraphics[width=1\linewidth]{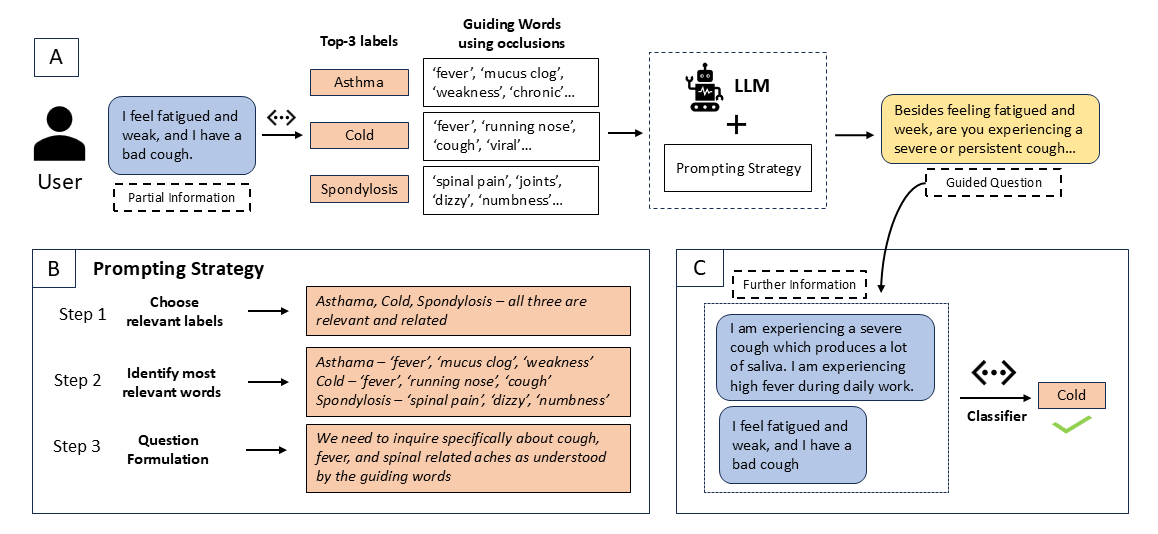}
    \caption{(A) Overall working framework of \model{} to leverage LLM and label explainability for asking guided question. (B) Details of the prompting strategy used. (C) Final classification along with incremental information.}
    \label{fig:enter-label}
\end{figure*}
Proactive questioning is another important limb of such systems wherein the bot engages with the user and directs the conversation ahead\cite{QA_classify, QA_IE}. It finds applications in many places like medical, customer support systems, and legal systems. 
%Previous works on Task Oriented Dialogues (ToD) also include maintaining a state of the conversation as a parameter.
%QA systems have been widely used in user text classification \cite{QA_classify} and proactive questioning for seeking further information~\cite{QA_IE} such as in dialog systems. 
A more specific challenge such systems face is of the static classification of user textual data. Often a textual data needs to be classified into a particular category or label \cite{classify_med,classify_legal}. For example, consider a system to classify a patient symptoms descriptions into one of the disease conditions, or a customer complaint system wherein a user writes a complaint to be categorized into a particular category. A practical challenge faced by such static categorization is inadequate or missing information toward appropriate routing of the user input to an actionable category. It can be immensely benefited by introduction of questioning component to prompt for further information or knowledge, grounded in previous response and the domain itself. Figure 1 shows an example of the same.

Our work introduces a novel framework, \model{}, aimed at framing guided questions based on prior partial information, such that specific relevant information can be asked for. This increment of information can aid for more accurate classification and completion of text. 
A classic techniques include classifier models for text classification\cite{BERT} \cite{ROBERTA} \cite{DEBERTA}. 
%Even though classifier models like DeBERTa \cite{DEBERTA} have shown impressive text classification abilities, the use of only a powerful classifier model may still be insufficient to compensate missing information. 
LLMs, on the other hand, have shown impressive abilities for reasoning and context understanding \cite{reason_LLm}. Further with techniques like in-context learning \cite{incon_intro} and chain of thought \cite{COT} with few-shot exemplars \cite{fewshot}, the task specific adaptation of LLMs greatly benefits. 
%The LLM performance also generally increases with parametric size of the models \cite{GPT4}. 
However, the LLM's parametric knowledge may still not capture domain specific requires. 
Finetuning options like FT (Full parametric training) and PEFT \cite{LORA} for learning from data history posses a huge challenge of high computation cost \cite{cost_LLM}.

Our framework, \model{} leverages the innate semantic understanding and reasoning ability of LLMs to combine with external classification explanability, for asking the most relevant guiding question. We specifically use LLaMa-3 8B model \cite{llama} for questioning and information seeking. 
We train BERT model \cite{BERT} for classification tasks on complete information which serves as the primary pivot for prediction of most probable labels. \model{} aims to use the inherent explanations of classifications for question formation.
For each class label we also learn the most significant words and phrases that contribute to the particular label classification. We use occlusions to find the keywords for a label using the training data.
The LLM utilizes the top most probable classifier labels and their significant representative keywords to form the guiding questions. 
Intuitively, the keywords are representative of the most important concepts present in the label knowledge. The summarized overview with keywords example can be found in figure 2.
%As an example, the most important keywords present in a label like 'fever' can be 'high temperature'; 'body ache'; 'yellow eyes', etc.
The LLM find the similar concepts present in the partial information and the labels, and frames the question based on the most distinguishing concepts between the label.
This help direct further information which may not be readily known earlier to the user. 
%As seen in Figure \ref{fig:intro_diag}, a patient initially provides an incomplete or partial information about their overall symptoms. The top-3 classifier labels are 'Asthma', 'Cold', and 'Spondylosis'. The bot frames a question asking about more specific relevant symptoms. Upon correct answering of the same, the extra information added helps to narrow between one of the three labels. 

We evaluate our framework, \model, on 8 text classification datasets. We first report the F1-Score of partial information and how accuracy changes by questioning and appending the new answer. We also show how explanability is influenced by keywords, namely by testing with unigrams, bigrams, and trigrams. We also report a higher question generation quality compared to other baseline methods.
Overall, our work, \model, presents a novel framework to ask guided questions when the initial text is incomplete or partial.
The question is such that it effectively differentiates between the most likely labels.

We summarize the key contributions of our work as follows: (i) We introduce a novel framework, \model{}, for providing guided questioning such that an initial partial information can be incremented. (ii) The guided questioning increments the information by leveraging explainability derived through top most confident classifier labels and their corresponding most significant key words. (iii) We show that further information collection through our framework significantly improves the classification accuracy as compared to other baselines. We also show that \model{} generates more accurate and targeted questions in relation to the user query.

\section{Related Works}
\subsection{Posthoc Explanations and LLMs}
Post-hoc explanations enhance the interpretability of Large Language Models (LLMs) by providing insights into which input features influence model outputs, addressing the "black-box" nature of these models \cite{posthoc_arellms}. The AMPLIFY framework \cite{posthoc_amplify}, for instance, uses attribution scores to generate natural language rationales, guiding LLMs to make more accurate predictions. Integrating LLMs with existing XAI algorithms like SHAP can produce more accessible and human-readable explanations. LLM-generated explanations are as effective as traditional gradient-based methods \cite{xai}.\\
Previous works use the technique of Integrated Gradients \cite{sundararajan2017axiomatic}, previously used for word-level attribution in language models\cite{enguehard2023sequential} to generate keywords that contribute the most for a particular classification label. 

\subsection{Task oriented dialogues}
Task-oriented dialogue systems aim to achieve specific goals through structured interactions, traditionally decomposing tasks into natural language understanding (NLU), dialogue management (DM), and natural language generation (NLG) . This modular approach often leads to error propagation and requires extensive domain-specific data \cite{tad_1}. Recent advancements, like SimpleTOD \cite{tad_simple}, leverage LLMs to unify NLU, DM, and NLG into a single sequence prediction task. By treating all sub-tasks as a single sequence prediction problem, SimpleTOD exploits transfer learning from pre-trained models for improved performance. Pre-trained generative models can effectively support ToD by learning domain-specific tokens.

\subsection{Information-seeking questions using LLMs}
Large Language Models (LLMs) have significantly advanced the field of information-seeking question answering by generating contextually rich and coherent responses \cite{info_chat}. Traditional search engines have evolved with models like GPT-3 and GPT-4, which can formulate search results in natural language, incorporating references from relevant sources. These models are particularly effective at summarizing and synthesizing information from various texts, making them valuable tools in medical and educational contexts. However, challenges such as "hallucination"—where the model generates plausible but incorrect information—highlight the need for mechanisms to verify and attribute sources accurately \cite{info_hag}. 
\subsection{In-context learning}
In-context learning (ICL) has emerged as a novel paradigm where language models are capable of learning from a few examples within a context, making predictions without explicit parameter updates. The key mechanism of ICL is the use of demonstration contexts, which consist of input-output examples formatted as natural language templates \cite{incon_gen}. This method allows large language models to perform various tasks by leveraging patterns learned from these demonstrations, which are provided as part of the input prompt. Recent studies highlight the adaptability of ICL to new tasks, significantly reducing computational costs compared to traditional supervised learning methods \cite{incon1}. Furthermore, ICL has demonstrated potential across different modalities, such as vision-language and speech tasks, by incorporating properly formatted data and architectural designs. Our work contributes to this growing field by utilizing ICL for the classification of incomplete sentences. By leveraging label explanations and guiding language models to ask targeted questions, we aim to refine user responses in medical contexts, ultimately enhancing the model's ability to generate accurate and relevant follow-up queries.
% \subsection{Integrated Gradients for post-hoc explainability}
% Integrated Gradients for 

\section{\model{} - Methodology}
\subsection{Overview}
% Consider a partial information provided initially, denoted as $x$, as the starting point. The aim is to classify the $x$, or $x$ with additional information into the correct category or label. Partial input refers to text that is incomplete, inadequate, or lacking critical details. We use a classifier model, $C$, to assign a text to one of the $n$ labels. The labels are dataset dependent and can be chosen from, say, the set $L = {l_1, l_2,..., l_n}$.
% We also identify the most significant keywords and phrases using occlusions associated with each label. For a label, $l_i$, we represent the set of keywords as  $r_i = {w^i_1, w^i_2,..., w^i_k}$. 

% \model{} considers a LLM with the following inputs: the top-k initial label predictions, their corresponding keywords, and a prompting strategy to leverage them and generate the guided question, $q$. In summary, the generated question focuses on the asking the user further information based on most relevant concepts in the keywords. This further helps to differentiate and classify into the correct label.

% In the following sections, we provide explanations for the detailed methodology used in our framework. The overall methodology is divided into three parts: (1) Classifier Training; (2) Learning Guiding words; and (3) Leveraging Explainability for Question Generation.

Given a partial input $x$, the objective is to classify $x$ (or $x$ with additional information) into the correct label. Here, partial input refers to text that may be incomplete or lacking essential details. We use a classifier, $C$, to map the input to one of $n$ labels, where the label set $L = \{l_1, l_2, \dots, l_n\}$ is dataset-dependent.
To enhance this process, we identify representative keywords and phrases using occlusion techniques for each label. For a label $l_i$, the associated keywords are denoted as $r_i = \{w^i_1, w^i_2, \dots, w^i_k\}$. 

\model{} leverages a large language model (Llama-3 8B in our study) to refine the classification through an interactive process. The model takes as input the top-$k$ predicted labels, their corresponding keywords, and a prompting strategy. It then generates a question $q$ designed to elicit information from the user related to the most relevant concepts in the keywords, further guiding the classification towards the correct label.

We divide our methodology as: (1) Classifier Finetuning, (2) Keyword Learning, and (3) Explainability-Driven Question Generation.
\begin{comment}
\begin{figure}
    \centering
    \includegraphics[width=0.9\linewidth]{latex/image-3.png}
    \caption{Enter Caption}
    \label{fig:enter-label}
\end{figure}
\subsection{Classifier Training}
\end{comment}
% The first step in our \model{} framework is the classifier training based on training data. The classifier model classifies a given text into a label, or in other words, provides the classification likelihood or probability for each label. 
% Our dataset comprises domain-specific texts paired with their corresponding labels, and we divide the data into training (80\%), evaluation (15\%), and test (5\%) splits.  We choose the BERT-uncased model as the classifier model, $C$. 
% Finetuning LLMs or more larger BERT-based models for classification task implies significantly increases computational cost. 
% The model is trained using the train split while the progress being evaluated the over the evaluation split.

% Importantly, we train using the complete text and its corresponding labels, in contrast to the partial information described earlier. As a result, the trained model outputs the most probable labels for a given input text.
\subsection{Classifier Finetuning}
The first step in our \model{} framework involves finetuning a classifier using labeled training data. 
%The goal of the classifier is to assign a probability distribution over labels for each input text. 
Our dataset consists of domain-specific text-label pairs, which are divided into training (80\%), evaluation (15\%), and test (5\%) sets. 
%We utilize the BERT-uncased model as the base classifier, $C$.

%While larger BERT-based models or fine-tuned LLMs could be employed, they come with a significant computational overhead.
We train $C$ on the complete input text and its corresponding label. Training is conducted on the training split, with performance monitored on the evaluation split.
By training on full texts, the model learns to output the most likely labels for any input, even if the input is incomplete during inference, as described in later sections.

\subsection{Keywords learning}
Given $n$ possible labels, $L = \{l_1, l_2, \dots, l_n\}$, the classifier assigns a probability score to each label for a given input. Our goal is to identify the most significant words or phrases (unigrams, bigrams, and trigrams) that represent each label. These keywords capture the core semantic and conceptual elements of the label’s category. For example, the label 'fever' might be characterized by keywords like "high temperature" or "body ache," while a telecommunication label might feature keywords like "mobile phone" or "network issues." These keywords enhance the explainability of the classification.

We employ the occlusion method to identify the top-$i$ significant words or word pairs for each label $l_i$, represented as $r_i = \{w^i_1, w^i_2, \dots, w^i_k\}$. Occlusions involve systematically removing or masking words from the input and observing the effect on the model’s confidence score. A sharp drop in confidence indicates that the removed word is crucial for the label prediction. Each word or phrase is then assigned a weight based on its importance, and the most relevant keywords are aggregated with additive weights for each label.

To capture diverse concepts, we consider unigrams, bigrams, and trigrams for each label. Figure 2(A) illustrates how these guiding keywords are used for question generation. In our experiments, we include the top 15 word pairs per label in the LLM prompt to guide the question generation process effectively.

\subsection{Explainability-Driven Question Generation}
%Starting with the partial input $x$, which may lack critical details for accurate classification, we generate a guided question to elicit additional information from the user, refining the classification process.
The final question generation stage follows the following method:
First, we pass the incomplete input $x$ to the classifier, which returns the top-$k$ labels with the highest confidence scores. We set $k = 3$, selecting the top-3 most likely labels for $x$. Using these labels, we prompt the LLaMA-3 8B model, applying a tailored prompting strategy to generate a guided question.

The prompt provided to the LLM includes the following components: (i) the partial input $x$; (ii) the top-3 predicted labels; (iii) the corresponding guiding keywords for each of these labels; (iv) a structured instruction prompt; and (v) a few-shot examples to guide the LLM’s output.

The generated question, $q$, can be formulated as:
\[
q = LLM(P \, || \, x \, || \, \{(l^x_1, r^x_1), (l^x_2, r^x_2), (l^x_3, r^x_3)\})
\]
where $P$ represents the instruction prompt, which combines the logical structure of the components with three few-shot examples. These examples illustrate how to generate questions that efficiently target missing or unclear information, aiding in more accurate classification.

The prompting strategy for generating questions follows a structured, three-step process:

\textbf{Step 1:} The LLM first filters out irrelevant labels from the top-3 predictions, retaining only those most relevant to the input query. This is achieved by comparing the input with the keywords associated with each label, allowing the model to focus on labels that contextually align with the partial information.

\textbf{Step 2:} Next, the LLM examines the guiding keywords for the remaining labels. These keywords represent key concepts associated with each label, serving as focal points for further inquiry. The LLM identifies the most contextually relevant keywords that can expand upon the incomplete information in the input $x$. This step ensures that the follow-up questions target the most meaningful aspects of the missing information.

\textbf{Step 3:} Finally, the LLM uses the selected keywords to generate a coherent, targeted question. The question is designed to elicit specific details necessary for distinguishing between the potential labels, facilitating more accurate classification. 

As shown in Figure 2(B), this strategy ensures a focused and contextually relevant interaction, especially for specialized categories, reducing the risk of hallucinations and irrelevant questions by grounding the response in the prior training data.
\begin{table*}[]
\centering
\resizebox{\textwidth}{!}{%
\begin{tabular}{@{}lcccccccc@{}}
\toprule
 & \multicolumn{4}{c}{BERT Classifier Model} & \multicolumn{4}{c}{DeBERTa Classifier Model} \\ \midrule
Dataset & partial & llm & LLM-nk & GuideQ & partial & llm & LLM-nk & GuideQ \\ \midrule
cnews & 45.2 & 48.5 (3.3) & 49.8 (4.9) & \textbf{50.9 (5.7)} & 42.6 & 45.5 (2.9) & 46.4 (3.8) & \textbf{49.6 (\underline{7.0})} \\
dbp & 86.9 & 87.0 (0.1) & 86.5 (-0.4) & \textbf{88.7 (1.8)} & 85.0 & 84.9 (-0.1) & 84.8 (-0.2) & \textbf{91.3 (\underline{6.3})} \\
s2d & 61.1 & 72.3 (11.2) & 66.9 (5.8) & \textbf{79.7 (18.6)} & 64.7 & 71.5 (6.8) & 68.4 (3.7) & \textbf{86.8 (\underline{22.1})} \\
salad & 35.2 & 53.6 (18.4) & 55.2 (20.0) & \textbf{57.1 (\underline{22.2})} & 38.0 & 55.7 (17.7) & 56.2 (18.2) & \textbf{58.7 (20.7)} \\
stress & 32.3 & \textbf{35.0 (2.7)} & 33.3 (1.0) & 32.9 (0.6) & 43.1 & 41.4 (-1.7) & 43.0 (-0.1) & \textbf{46.1 (\underline{3.0})} \\
20NG & 67.5 & 68.2 (0.7) & 68.0 (0.5) & \textbf{72.9 (\underline{5.4})} & 63.2 & 64.0 (0.8) & 63.9 (0.7) & \textbf{65.8 (2.6)} \\ \bottomrule
\end{tabular}%
}
\caption{Comparison of \% F1-Scores of \model{} along with three baseline approaches - (i) partial: partial information; (ii) LLM: Only LLM is used for question framing; (iii) LLM-nk: LLM is provided with top 3 predictions. The results are reported for two classifier models: BERT and DeBERTa. Numbers in bracket constitute gain over partial information scores}
\label{tab:table1}
\end{table*}

% \section{Experimental Setup}
% In this section, we describe the various experimental settings and evaluation methods used to test our framework. The primary objective of our framework, \model, is to enhance classification of partial or incomplete information by generating guided questions which prompt for additional, relevant information. This helps in a more accurate classification of the texts by targeted filling of essential missing information, grounded in previous training data.

% \subsection{Datasets}
% We evaluated our \model{} framework on three text classification datasets: (i) Symptom2Disease, (ii) Crypto News, and (iii) Human Stress Prediction to capture performance across multiple domains. The Symptom2Disease dataset is critical for testing our framework in healthcare, where accurate classification based on partial symptoms is vital. The Crypto News dataset allows us to explore how our approach handles dynamic and rapidly changing information, crucial in financial contexts. The Human Stress Prediction dataset is particularly relevant for assessing our framework's ability to improve predictions in the psychological and behavioral domain, where information is often incomplete. We have divided these datasets into 80\% for training, 15\% for validation, and 5\% for testing to ensure a balanced approach that optimizes model learning while providing a robust evaluation of unseen data. This strategy ensures that our framework is rigorously tested and performs effectively in real-world scenarios across various applications.
\section{Experimental Setup}
This section outlines the experimental configurations and evaluation protocols employed to assess the performance of our framework, \model. The goal of \model{} is to enhance classification under incomplete information by generating guided questions that elicit relevant, missing inputs. This targeted information retrieval facilitates more accurate classification, grounded in prior learned patterns.

\subsection{Datasets}
We evaluated \model{} on six diverse text classification datasets, each representing a unique domain and presenting distinct challenges: (i) Symptom2Disease, (ii) Crypto News, (iii) Human Stress Prediction, (iv) 20 Newsgroups, (v) DBpedia, and (vi) SALAD-Bench, representing healthcare, financial, and behavioral domains respectively. The Symptom2Disease dataset challenges the model’s ability to classify diseases based on incomplete symptom descriptions, critical in medical decision-making. Crypto News assesses the framework’s adaptability to rapidly evolving financial data. Human Stress Prediction evaluates the model in a psychological context, where input data is often sparse or incomplete. The 20 Newsgroups dataset, with its wide range of discussion topics, tests the model's generalizability across various themes. DBpedia challenges the model with a broad spectrum of structured factual information, essential for handling real-world knowledge-based queries. Finally, we introduce SALAD-Bench, a novel benchmark designed for evaluating LLMs in safety, defense, and attack scenarios.We split each dataset into 80\% training, 15\% validation, and 5\% test sets, ensuring a rigorous evaluation. This setup offers a comprehensive assessment across varied real-world applications.

% \subsection{Baselines}
% We compare \model{} with three additional baselines for question generation. (i) We use the partial information originally provided as the input as one of the baselines. It signifies the classification ability using just the partial information. (ii) We use only LLM as the question generator as the second baseline, \textit{i.e.}, the LLM (Llama-3 8B) is provided with the partial input and prompted to ask a question to complete the missing information. As the input provided is a semantic context related to a realistic setup as opposed to a fictitious or made-up content, the LLMs can also question over it. (iii) As a third we use LLM (Llama-3 8B) along with the top-3 classifiers. The LLM is prompted to generate question based on the top-3 likely classifications, without the keywords. We choose this to compare the effects of explanability brought by only labels, as opposed to keywords and labels in \model{}. We refer the three baselines as $partial$, $LLM$, and $LLM+labels$ respectively.
\subsection{Baselines}
We benchmark \model{} against three baselines for question generation: (i) \textit{Partial}: using only the initially provided partial information to assess classification performance without any additional inputs, serving as a direct comparison of classification under incomplete data. (ii) \textit{LLM}: leveraging a standalone LLM (Llama-3 8B), prompted to generate questions based on the partial input, representing a generic approach to eliciting missing information. As the input data mirrors realistic scenarios, the LLM's questions are grounded in semantic context and not in hypothetical constructs. (iii) \textit{LLM-nk}: LLM with only labels and no keywords - combining the LLM with the top-3 predicted classifications, where the LLM generates questions based solely on these labels without keywords. These baselines, denoted as Partial, LLM, and LLM-nk, provide a comprehensive evaluation of \model's performance.

% \usepackage{booktabs}
% \usepackage{graphicx}

% \subsection{Evaluation}
% We for various experiments to evaluate different aspects of the problem. We finetune BERT-uncased model on each dataset separately for classification task. Llama-3 8B, an open-source LLM model, is used throughout our experiments. We prefer a comparable smaller parametric model due to it's significant saving of computational costs. Llama-3 8B model is known to have good reasoning ability as well as a comparatively smaller size. 
% For testing, each data instance is split into two concurrent equal parts. The first part acts as the partial information. We use the second half acts as the reference for the generated question to answer from. We show our results using RoBERTa and DistillBERT as the question answering model, which extracts the most relevant text from the reference text in relation to the question asked. 
\subsection{Evaluation}
We conduct multiple experiments to assess various aspects of the problem. For a robust classification analysis, we use two classifier models: BERT-uncased and DeBERTaV3 (a comparatively larger model). Throughout our experiments, we utilize Llama-3 8B, an open-source LLM chosen for its strong reasoning capabilities and computational efficiency, offering a balance between performance and cost due to its smaller parametric size.

During testing, each data instance is split into two equal parts: the first half serves as the partial input, while the second half acts as the reference from which the generated question seeks to extract missing information. To evaluate question quality and answer relevance, we employ DeBERTaV3 finetuned on SQuadV2, as the question-answering models, which extract the most relevant text from the reference based on the question. We set a 20\% confidence threshold for answering. 

% We demonstrate four experimental settings for analysis of \model{}. 
% \\(i) \textbf{Classification Accuracy} We report the classification accuracy, \textit{i.e.}, the accuracy for correct classification of a given text to the right label. We first report the scores on partial information (first half of an instance). For other baselines, a question is constructed and the previously mentioned QA models extract relevant phrases from the reference text. The extracted text is appended to the partial information, and the combined text is considered for classification.
% \\ (ii) \textbf{Question Quality} Next, we evaluate the quality and relevance of the questions generated themselves through different baselines. We consider win rate between pairs of methods, \textit{i.e.}, choosing the most relevant among the two in relation to complete text (both partial input and reference together). We report the win rate accuracy for each pair. The pairs are as follows: $(\model{}, LLM)$, $(\model{}, LLM+labels)$, $(LLM+labels, LLM)$. 
% \\ (iii) \textbf{Explanability by keywords} We use three variations of keywords corresponding to each label - unigram, bigram, and trigram. We compare the results of \model{} using 
% \\ (iv) \textbf{Multi Turn}

We demonstrate four experimental settings to thoroughly analyze \model{}:
\\
\textbf{(i) Classification Performance:} We measure classification performance, which refers to correctly assigning the given text to its corresponding label by reporting F1-Scores. First, we report scores using only the partial information (the first half of each instance). For the other baselines, a question is generated, and relevant phrases are extracted from the reference text using the QA models with a 20\% confidence threshold. This extracted information is appended to the partial input, and the combined text is classified.
\\
\textbf{(ii) Question Quality:} We calculate the win rate between pairs of methods by determining which generated question is more aligned with the complete text (both partial input and reference). The win rate accuracy is reported for the following pairs: $(\model{}, LLM)$, and $(\model{}, LLM+Labels)$.
\\
\textbf{(iii) Explainability via Keywords:} We explore the explainability provided by different keyword types associated with each label—unigram, bigram, and trigram. This analysis compares the impact of keyword granularity on \model's performance, helping us understand how keyword-based question generation improves the classification process.
\\
\textbf{(iv) Multi-Turn Interaction:} We investigate the potential of multi-turn question generation, where multiple rounds of guided questions are used to iteratively refine the extracted information. This setting evaluates how effectively \model{} handles scenarios where a single question is insufficient to gather all relevant details.

% \subsection{Experiments}
% \textbf{Setup:} We finetune BERT-uncased model as the classifier model. Further, we use the Llama-3 8B model as the LLM for question generation in our framework. We prefer a comparably smaller parametric model due to lower computational requirements. For a given dataset, we perform a sentence-level split for the text instances by dividing into two equal parts. The first half of the text acts as the partial information for the user. We call the second half as 'grounded context', which is used to extract answers for the generated question. We use RoBERTa and DistillBERT as Question Answering model. The models extract the answer snippets from the grounded contexts. 
% \usepackage{booktabs}
% \usepackage{graphicx}
% \usepackage{booktabs}
% \usepackage{graphicx}

% \usepackage{booktabs}
% \usepackage{graphicx}
\textbf{Setup:} We fine-tune a BERT-uncased model as the classifier in our framework. Additionally, we use the Llama-3 8B model for question generation. We select this smaller parametric model to reduce computational overhead while maintaining robust performance. For each dataset, we split the text instances at the sentence level, dividing them into two equal parts. The first half serves as the partial information provided to the model. The second half, referred to as the "grounded context", is used to extract answers to the generated questions. RoBERTa and DistillBERT are employed as the Question Answering models, tasked with extracting the most relevant answer snippets from the grounded contexts.

\begin{table}[]
\centering
\resizebox{\columnwidth}{!}{%
\begin{tabular}{@{}lcclcc@{}}
\toprule
\multicolumn{6}{c}{Skyline Classification} \\ \midrule
Datasets & \begin{tabular}[c]{@{}c@{}}F1-score/\\ BERT\end{tabular} & \multicolumn{1}{c|}{\begin{tabular}[c]{@{}c@{}}F1-score/\\ DeBERTa\end{tabular}} & Datasets & \begin{tabular}[c]{@{}c@{}}F1-score/\\ BERT\end{tabular} & \begin{tabular}[c]{@{}c@{}}F1-score/\\ DeBERTa\end{tabular} \\ \midrule
cnews & 63.8 & \multicolumn{1}{c|}{65.6} & salad & 64.5 & 66.4 \\
dbp & 95.5 & \multicolumn{1}{c|}{94.5} & stress & 43.5 & 47.7 \\
s2d & 99.8 & \multicolumn{1}{c|}{100.0} & 20NG & 75.8 & 71.6 \\ \bottomrule
\end{tabular}%
}
\caption{Skyline F1-Scores for complete original text (both partial and reference combined) on BERT and DeBERTa classifier models}
\label{tab:skyline}
\end{table}
\section{Results and Observations}
Our work, \model{} involved an evaluation using six classification datasets, which allowed us to rigorously test the framework's ability to generate guided questions based on partial information and enhance classification accuracy.

The strongest results of GuideQ framework can be seen in table 1, wherein we report the F1 score for classification task post-answering the generated question. The question answering model is such that it extracts the most relevant text from the reference text in respect to the asked question. We observe that for all datasets at all instances (expect for stress with BERT classifier), our method shows the highest overall classification scores (F1 score) across both classifier models. 
Second, considering results of both classifier models together, table 1 also shows that \model{} always has the highest margin of improvement compared to F1 scores with partial information. 

We observe an improvement of 18.6\% (BERT) and 22.1\% (DeBERTa) for S2D and 22.1\% (BERT) and 20.7\% (DeBERTa) for SALAD dataset with our method over partial information. This large shift shows our method is effectively able to frame questions based on previous training data and keywords. This makes the question asked more grounded in context of the partial information. 
A question maybe raised as to why some datasets show more improvement in F1 score than the others.
We observe the skyline F1 score results for each dataset using the complete text on both classifier models (table 2). The skyline results reveal the inherent classification ability of the datasets themselves.

We also observe that in some situations with other baselines, the addition of new answer based on generated question negatively impacts the classification, \textit{i.e.}, reduces F1 score. Example of the same is a drop of 0.4\% for DBP dataset with BERT classifier and a drop of 0.2\% with DeBERTa classifier using LLM-nk approach. Many instances also include when the baselines show only a slight improvement. However, our method never shows a dip in score compared to partial information classification. The range of percentage gain is also comparatively larger than other methods. This is crucial in realistic situations where we want to ensure a originally correct categorized text is not misclassified.
\section{Analysis}
\subsection{Effect of Classifier Model}
We perform our experiments by finetuning two different classifier models - BERT and DeBERTa for a robust analysis. Firstly, \model{} outperforms other methods on both the classifier models in most settings. Only for stress dataset with BERT classifier, a classic LLM based approached worked better and showed 2.7\% increase in F1 score.
When absolute percentage gain over partial information is taken into consideration for our method, classification using DeBERTa leads for four out of six datasets. BERT shows highest absolute gain for salad and 20NG datasets. We represent the percentage gains in brackets in table 1 and underline the highest gain across a dataset. Although we observe comparable gains for partial information classification along with LLM, and LLM-nk baseline approaches. 

A possible reason for the same maybe as follows: DeBERTa being a larger parameter model compared to BERT is able to capture more relevant explainable keywords for a given label using occlusions. This results in a better formed and focused question. The conclusion we can derive is that even though a better classifier model doesn't necessarily show higher F1 scores with partial information, it aids to improve results when combined with \model{} framework leveraging explainable keywords.
% \usepackage{booktabs}
% \usepackage{graphicx}
\begin{table}[]
\centering
\resizebox{\columnwidth}{!}{%
\begin{tabular}{@{}lrrlrr@{}}
\toprule
\multicolumn{6}{c}{Win Rate Scores} \\ \midrule
Datasets & \multicolumn{1}{c}{LLM} & \multicolumn{1}{c|}{LLM-nk} & Datasets & \multicolumn{1}{c}{LLM} & \multicolumn{1}{c}{LLM-nk} \\ \midrule
cnews & 66.0\% & \multicolumn{1}{r|}{67.0\%} & salad & 72.0\% & 62.0\% \\
dbp & 93.0\% & \multicolumn{1}{r|}{92.0\%} & stress & 65.0\% & 70.0\% \\
s2d & 90.0\% & \multicolumn{1}{r|}{85.0\%} & 20NG & 89.0\% & 93.0\% \\ \bottomrule
\end{tabular}%
}
\caption{Win Rate (WR) \% scores of questions generated for (i) \model{} with LLM baseline; (ii) \model{} with LLM-nk baseline}
\label{tab:win_rate}
\end{table}

\begin{table}[]
\centering
\resizebox{0.86\columnwidth}{!}{%
\begin{tabular}{@{}lccc|ccc@{}}
\toprule
 & \multicolumn{3}{c|}{BERT} & \multicolumn{3}{c}{DeBERTa} \\ \midrule
Dataset & uni & bi & tri & uni & bi & tri \\ \midrule
cnews & \textbf{6.8} & 5.6 & \underline{6.1} & \underline{7.2} & 7.1 & \textbf{8.9} \\
dbp & \textbf{2.5} & \underline{1.8} & 1.6 & \textbf{6.6} & \underline{6.3} & 6.2 \\
s2d & \textbf{19.7} & \underline{18.6} & 17.0 & \underline{14.6} & \textbf{22.1} & 12.8 \\
salad & 20.7 & \textbf{21.9} & \underline{21.3} & 20.1 & \textbf{20.7} & \underline{20.6} \\
stress & -0.3 & \underline{0.6} & \textbf{1.6} & \underline{-0.2} & \textbf{3.0} & -1.5 \\
20NG & \textbf{5.8} & \underline{5.4} & 5.0 & \textbf{3.9} & 2.7 & \underline{3.0} \\ \bottomrule
\end{tabular}%
}
\caption{Comparison of \% absolute gain over partial information F1-Scores for \model{} framework with unigram (uni), bigram (bi), and trigram (tri) keywords }
\label{tab:table3}
\end{table}

\subsection{Analysis of baseline approaches}
We also observe that the performance of the other two baselines, \textit{i.e.}, the use of only an LLM to ask relevant question and providing an LLM with only the top-3 classifier labels perform almost at par with each other. In other words, though we conclude that \model{} shows improved performance over other baselines, the two baselines themselves are comparable to each other in performance.
For example, for SALAD dataset LLM-nk baseline performs better while for S2D dataset only LLM baseline shows higher results. This holds true for both classifier models. 
While \model{} leverages explainability through keywords along with labels, the LLM-nk baseline uses only labels. Apparently, the labels themselves do not add explicit information that would help guide for completion of the text.

\subsection{Quality of Generated Questions}
Next, we evaluate the question quality of different baselines. We do this by calculating the win rates using LLM model of question generated with our method taking LLM and LLM-nk as base on a subset of 100 random instances for each dataset. Table 3 summarizes the results for the same. We observe that our method always has a win rate above 50\%. The minimum win rate reached is 62.0\% for salad datasets with LLM-nk. For three datasets: db, s2d, and 20NG, \model{} performs exceptionally higher. Overall, table 3 shows that our method frames questions which are more relevant and specific to the partial information and unseen reference answer.

\subsection{Effect of n-grams in \model{}}
In this section we compare the results three different n-gram approaches for keywords generation, namely: unigram, bigram, and trigram, which means the keywords are restricted exactly to be single words, two words, and three words respectively. The results comprising of percentage gain of F1 score over that of partial information are summarized in table 3. We observe that while all the three perform almost similar, still results with questions formed using unigram and bigram are superior in 30/36 total setups as compared to results using trigram. In other words, \model{} slightly performs better when the explainable keywords are restricted to two or three words. Although there is no particular trend between unigram and bigram themselves. 

\section{Ablation Study: Multi-turn}
Our framework explores a multi-turn setting, where successive guided questions are posed following a prior response. We specifically report multi-turn results on the CNEWS dataset due to its larger context per instance and significant differences in partial information scores versus skyline F1-Scores. The text is divided into three segments, with the first serving as partial information to initiate the guided questioning (GuideQ) process. This method involves dynamically updating the pool of guiding words, removing those already used in previous turns. After generating questions, the answer extraction model derives responses from the text. This answer, combined with the initial partial information and the refreshed guiding words, informs the next GuideQ round. This cycle repeats over three turns, continually refining the guiding words to enhance the relevance and depth of information retrieval. Figure 3 illustrates the summarized results, showing our method's superior performance in multi-turn scenarios.
\begin{figure}
    \centering
    \includegraphics[width=\linewidth]{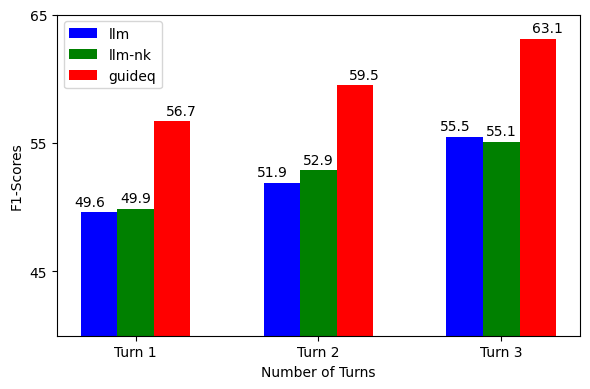}
    \caption{Multiturn Results: F1-Scores for three turn question answering on CNEWS datatset.}
    \label{fig:enter-label}
\end{figure}

\section{Conclusion and Future Work}
Our work \model{}, presents a novel framework for generating guided questions to enhance classification of partial information. By leveraging explainable keywords derived from classifier models along with LLM-based question generation, GUIDEQ demonstrates superior performance across multiple datasets compared to baseline approaches. Our results show consistent improvements in F1 scores, with gains of up to 22\% on certain datasets. \model{} generate high-quality, context-relevant questions is evident from the win rates against baseline methods. \model{}'s effectiveness in multi-turn interactions and its flexibility in accommodating different n-gram approaches for keyword generation further underscore its potential for real-world applications in information retrieval and classification tasks.

\section{Limitations}
Despite GUIDEQ's promising results, several limitations should be noted. First, the framework's performance is dependent on the quality of the initial classifier model and the relevance of extracted keywords. Suboptimal classifier training or keyword selection could lead to less effective question generation. Secondly, the  framework's reliance on LLMs for question generation also introduces potential biases and inconsistencies inherent to these models. Finally, the computational resources required for running large language models may pose scalability challenges in certain applications.

\section{Acknowledgments}
We are thankful for the support provided by BharatGen for analytics and compute

%\section{Appendices}

\bibliography{custom}

\appendix

\section{Further Results of classification}
We further report the recall and precision scores for the main classification task following table 1 results. The recall and precision values show a similar trend as that of F1-score in table 1. We report precision scores in \ref{tab:precision} and recall in \ref{tab:recall}.
% \usepackage{booktabs}
% \usepackage{graphicx}
\begin{table*}[]
\centering
\resizebox{0.9\textwidth}{!}{%
\begin{tabular}{@{}lrrrrrrrr@{}}
\toprule
 & \multicolumn{4}{l}{BERT Classifier Model} & \multicolumn{4}{l}{DeBERTa Classifier Model} \\ \midrule
Dataset & \multicolumn{1}{l}{partial} & \multicolumn{1}{l}{llm} & \multicolumn{1}{l}{LLM-nk} & \multicolumn{1}{l}{GuideQ} & \multicolumn{1}{l}{partial} & \multicolumn{1}{l}{llm} & \multicolumn{1}{l}{LLM-nk} & \multicolumn{1}{l}{GuideQ} \\
cnews & 0.489 & 0.514 & 0.523 & 0.545 & 0.568 & 0.566 & 0.574 & 0.603 \\
dbp & 0.943 & 0.934 & 0.926 & 0.913 & 0.926 & 0.919 & 0.919 & 0.932 \\
s2d & 0.625 & 0.757 & 0.722 & 0.805 & 0.697 & 0.763 & 0.757 & 0.793 \\
salad & 0.414 & 0.584 & 0.607 & 0.622 & 0.447 & 0.613 & 0.611 & 0.627 \\
stress & 0.323 & 0.352 & 0.338 & 0.330 & 0.471 & 0.426 & 0.468 & 0.421 \\
20NG & 0.697 & 0.699 & 0.700 & 0.748 & 0.659 & 0.674 & 0.688 & 0.704 \\ \bottomrule
\end{tabular}%
}
\caption{Precision}
\label{tab:precision}
\end{table*}

% \usepackage{booktabs}
% \usepackage{graphicx}
\begin{table*}[]
\centering
\resizebox{0.9\textwidth}{!}{%
\begin{tabular}{@{}lrrrrrrrr@{}}
\toprule
 & \multicolumn{4}{l}{BERT Classifier Model} & \multicolumn{4}{l}{DeBERTa Classifier Model} \\ \midrule
Dataset & \multicolumn{1}{l}{partial} & \multicolumn{1}{l}{llm} & \multicolumn{1}{l}{LLM-nk} & \multicolumn{1}{l}{GuideQ} & \multicolumn{1}{l}{partial} & \multicolumn{1}{l}{llm} & \multicolumn{1}{l}{LLM-nk} & \multicolumn{1}{l}{GuideQ} \\
cnews & 0.436 & 0.469 & 0.484 & 0.496 & 0.418 & 0.440 & 0.451 & 0.495 \\
dbp & 0.838 & 0.846 & 0.845 & 0.885 & 0.820 & 0.825 & 0.825 & 0.912 \\
s2d & 0.653 & 0.729 & 0.681 & 0.799 & 0.667 & 0.722 & 0.688 & 0.806 \\
salad & 0.358 & 0.547 & 0.562 & 0.571 & 0.374 & 0.556 & 0.561 & 0.591 \\
stress & 0.339 & 0.364 & 0.347 & 0.368 & 0.427 & 0.417 & 0.424 & 0.433 \\
20NG & 0.678 & 0.688 & 0.685 & 0.730 & 0.641 & 0.649 & 0.648 & 0.679 \\ \bottomrule
\end{tabular}%
}
\caption{Recall}
\label{tab:recall}
\end{table*}

\label{sec:appendix}
\section{Prompts used in our experiments}
\subsection{Prompt for LLM-nk baseline}

You are a AI expert. You are provided with partial medical information along with the top-3 label where the information can belong to.  Your task is to ask an information-seeking question based on the partial information and the labels such that when answered, one of the labels can be selected with confidence.

        Follow the following thinking strategy:
        First, eliminate the labels that are not probable for the given information. Identify the main context of the partial information and see if a similar content matches in any of the  label. If it doesn't, then the label can be taken out of consideration.
        Now generate a question. This question should further probe for information that will help refine the identification of the most likely diagnostic category. 

        Generate both the thought process and the question. Strictly follow the format shown in examples for output generation. Double quote the final question.

        Here are a few examples to understand better:

        *

        Example 1:
            Partial information: I constantly sneeze and have a dry cough.

            Category: Allergy 
            Category: Diabetes 
            Category: Common Cold 

            QUESTION: "Besides fever, are you experiencing symptoms such as cough, severe headaches, localized pain, or inflammation? Also, can you describe the pattern of your fever—is it continuous or does it occur in intervals?"

        Example 2:
            Partial information: The software keeps crashing.

            Category: Software Bug 
            Category: User Error 
            Category: Hardware Issue 

            QUESTION: "When the software crashes, do you receive any specific error messages, or does it happen during particular tasks? Have you noticed any hardware malfunctions or overheating before the crashes?"

        Example 3:
            Partial information: The car is making a strange noise.

            Category: Engine Problem 
            Category: Tire Issue 
            Category: Transmission Issue 

            QUESTION: "Can you describe the noise in more detail? Is it a grinding, squealing, or clicking sound? Does it happen while driving, when shifting gears, or when the car is stationary?"

            *
            
            Now generate note and QUESTION for:
                
            """

\section{Prompt for LLM baseline}
input text = """
 
    You are an AI Expert. You are provided with partial information along with the top-3 categories where this information could belong. Each category also has a list of keywords that represent the characteristic content covered by the category. Your task is to ask an information-seeking question based on the partial information and the category keywords such that when answered, one of the categories can be selected with confidence.

    Follow the following thinking strategy:
    First, eliminate the categories that are not probable based on the given information. Identify the main context of the partial information and see if similar content matches any of the keywords in a category. If it doesn't, then the category can be taken out of c…
You are a AI expert. You are provided with partial   information.  Your task is to ask an information-seeking question based on the partial information  such that when answered, the classification of the labels can be done with confidence.

    Follow the following thinking strategy:
     Identify the main context of the partial information now generate a question. This question should further probe for information that will help refine the identification of the most likely diagnostic category. 

    Generate both the thought process and the question. Strictly follow the format shown in examples for output generation. Double quote the final question.

    Here are a few examples to understand better:

    *

    Example 1:
        Partial information: I constantly sneeze and have a dry cough.
 
        QUESTION: "Besides fever, are you experiencing symptoms such as cough, severe headaches, localized pain, or inflammation? Also, can you describe the pattern of your fever—is it continuous or does it occur in intervals?"

        Example 2:
        Partial information: The software keeps crashing.

        QUESTION: "When the software crashes, do you receive any specific error messages, or does it happen during particular tasks? Have you noticed any hardware malfunctions or overheating before the crashes?"

        Example 3:
        Partial information: The car is making a strange noise.

        QUESTION: "Can you describe the noise in more detail? Is it a grinding, squealing, or clicking sound? Does it happen while driving, when shifting gears, or when the car is stationary?"

        *
        
        Now generate note and QUESTION for:
            
        """
\end{document}